# Assessment of Amazon Comprehend Medical: Medication Information Extraction

**Benedict Guzman, MS, Isabel Metzger, MS, Yindalon Aphinyanaphongs, M.D., Ph.D., Himanshu Grover, Ph.D.**
NYU Langone Health, New York, NY

**Abstract**
*In November 27, 2018, Amazon Web Services (AWS) released Amazon Comprehend Medical (ACM), a deep learning-based system that automatically extracts clinical concepts—which include, anatomy, medical conditions, protected health information (PHI), test names, treatment names, and medical procedures, and medications—from clinical text notes. Uptake and trust in any new data product relies on independent validation across benchmark datasets and/or tools to establish and confirm expected quality of results. This work focuses on the medication extraction task, and particularly, ACM was evaluated using the official test sets from the 2009 i2b2 Medication Extraction Challenge and 2018 n2c2 Track 2: Adverse Drug Events and Medication Extraction in EHRs. Overall, ACM achieved F-scores of 0.768 and 0.828. These scores ranked the lowest when compared to the three best systems in the respective challenges. To further establish generalizability of ACM's medication extraction performance, a set of random internal clinical text notes from NYU Langone Medical Center were also included in this work. And in this corpus, ACM garnered an F-score of 0.753.*

## I. Introduction

With the institution of the Health Information Technology for Economic and Clinical Health Act (HITECH) in 2009, there has been a rapid increase in the adoption of electronic health records (EHRs) in healthcare institutions[1]. Majority of data in EHRs are in the form of free text, which is a more natural and expressive medium for documentation and communication[2]. Clinical text notes feature a gold mine of information such as detailed patient conditions, treatment strategy, and prescribed medications. However, due to its unstructured nature, the information from these notes must be extracted and categorized to be utilized for clinical decision support, quality improvement, and research[2]. Employing domain experts to perform the task manually is one solution, but this can be time-consuming and error-prone[3]. Therefore, there is a necessity for automated systems that could parse medical information with high efficiency and accuracy.

A feasible solution is natural language processing (NLP) techniques. Specifically, its subdomain, information extraction (IE) aims to automatically output structured information from unstructured documents[4]. Within the field of IE, there are multiple subtasks, and two of which are: named entity recognition (NER), which deals with recognition of entities and classification into predefined categories such as person names, places, or organizations, and relationship extraction (RE), which focuses on identification of the relationships amongst extracted entities[5-6].

There are mainly two types of methods employed for these subtasks. The first and the predominant approach used in the clinical domain is rule-based, which consists of a collection of handcrafted rules—such as regular expressions and logic—that require collaboration with domain experts[2]. The second method is machine learning. Although underutilized in the clinical domain, this approach is more portable and scalable, and overall, produces better results provided that the model is trained using a large dataset[7]. Some successful systems, however, utilize both methods concurrently, and thus are called hybrid systems[8].

Several technological companies have developed their own IE systems. Most recently, Amazon Web Services (AWS) launched Amazon Comprehend Medical (ACM). ACM's machinery is driven by state-of-the-art deep learning models[9]. The underlying algorithms and technology that power ACM have already been implemented and trained and are updated routinely as end-user requirements evolve. Multiple modes of access—web console and programmatic access via Python or Java—cater to a wide audience with varying levels of skill and facility for technology. Users only pay for what they use, with no minimum fees and upfront commitments (i.e. $0.01 per unit, where 1 unit = 100 UTF-8 characters)[9]. Currently, only clinical text notes written in English are supported by ACM[9].

ACM automatically identifies entities, relationships, and negation of concepts (e.g. patient did not take the medication due to known allergic reactions) via the protected Medical Protected Health Information Data Extraction and Identification (PHId), which focuses on protected health information (PHI) only, and Medical Named Entity and Relationship Extraction (NERe) API, which detect the following clinical concepts:

1. Anatomy: relates to the body parts and systems and the corresponding location/directionality (e.g. dorsal, ventral, proximal, and distal).
2. Medical Condition: involves the diagnosis name and the corresponding acuity (e.g. chronic or acute), signs and symptoms.
3. Protected Health Information: focuses on various PHI such as patient's name, birthdate, and social security number.
4. Test Name, Treatment Name, and Procedure Name: deals with diagnostic testing, interventions, and treatment procedures that relate to a medical condition.
5. Medications: includes the name (i.e. generic name and brand name), dosage, duration, frequency, form, frequency, mode, rate, and strength.

An example ACM output is provided in figure 1. Although not shown in figure 1, each of its output has a corresponding score that indicates the confidence level (scale from 0.0 to 1.0, with 1.0 as the highest confidence level) for the extraction accuracy.

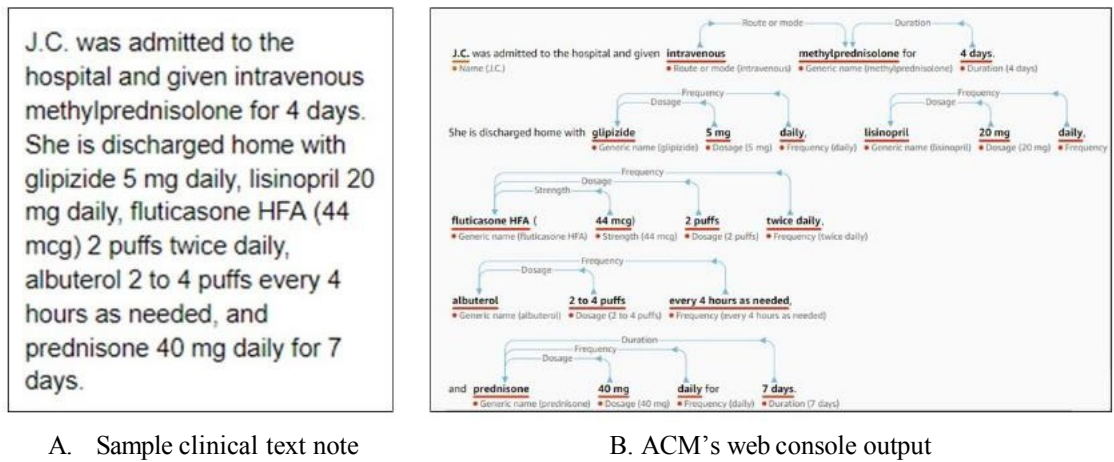

A. Sample clinical text note    B. ACM's web console output

**Figure 1**: (A) Sample clinical text note as adapted from Koda Kimble and Young's Applied Therapeutics - The Clinical Use of Drugs: Case 102-5[10] and (B) ACM's web console output, with extracted medical entities and their relationships[9].

Uptake and trust in any new data product—especially the ones applied to delicate medical information—relies on independent validation across benchmark datasets and/or tools to establish and confirm expected quality of results. This work focuses on medication extraction, which seeks to draw out medication entities and relationships from clinical notes. Entities include attributes such as name, dosage, frequency, mode, duration, strength, and form. Accurate retrieval and dissemination of these information is essential because medications play a vital role in patient care, especially in disease prognosis and survival. Medication errors could lead to adverse drug events, which is any harm or injury experienced by the patient due to medication-related intervention[11]. Such cases are attributed to 7,000-9,000 preventable deaths annually, and associated expenses by the US government total to $40 billion [12].

Specifically, this work has the following objectives:
1. ACM's first version (which was released in November 27, 2018) was evaluated using the official test sets of the 2009 Informatics for Integrating Biology & the Bedside (i2b2) Medication Extraction Challenge and 2018 National NLP Clinical Challenges (n2c2) Track 2: Adverse Drug Events and Medication Extraction in EHRs for medication extraction. Specifically, the quality of ACM's medication information extraction outputs was examined by comparing against gold standard (ground truth labels) annotations from the corresponding datasets.
2. For comparative analysis, ACM's performance was contrasted against the three best systems in the respective challenges. For the i2b2 challenge, these include systems by Tao et. al, Patrick et. al, and Doan et. al, and for the n2c2 challenge, these include systems by Alibaba Inc., UTHealth/Dalian, and University of Florida.
3. To further establish generalizability of ACM's medication extraction performance, internal random clinical notes from NYU Langone Medical Center were also included in the evaluation.

## II. Amazon Comprehend Medical and Related Work

Amazon Comprehend Medical

ACM's is backed by deep learning algorithms. To perform NER, it utilizes a neural network architecture which constitutes two Long Short-Term Memory (LSTM) encoders at the character and word level, and a single LSTM tag decoder[13]. This base framework is further modified by the addition of transfer learning via Tunable Transfer network (TTN) and Dynamic Transfer Networks (DTN), which exploit parameter sharing schemes and gating[14]. Such configuration overcomes the major limitation of limited access to medical data for training purposes and ensures generalizability of the system across different medical specialties[14]. Outputs from the NER decoder are fed to an entity softmax layer, which then are used as input for the negation softmax layer to detect negation of concepts[15]. Finally, ACM's RE function features a method that combines second-order relation scores—which utilizes a context token that connects the two target entities—and first-order relation scores[16].

2009 i2b2 Medication Extraction Challenge Three Best Systems

20 teams representing 23 organizations and 9 countries participated in the original challenge; 10 systems employed a rule-based approach, 4 used a machine learning-based approach, and 6 used a hybrid approach[17]. The best performing system in the challenge by Patrick et. al is based on hybrid of conditional random fields (CRFs) for the NER task and support vector machines for the RE task[18]. Doan et. al's system garnered the second-best performance in the challenge, and their algorithm is an extension of MedEx by Xu et.al, which relies purely on rules such as lookup, regular expression, and rule-based disambiguation components[19]. Although not originally a participant of the 2009 challenge, a more recent machine learning algorithm by Tao et. al (2017) was trained and tested using the i2b2 dataset. Their system features word embeddings and CRFs to extract medication entities and its attributes[20].

2018 Track 2: Adverse Drug Events and Medication Extraction in EHRs Three Best Systems

The pool of participants for the n2c2 challenge was composed of 28 teams, which represented 11 countries[21]. The top 10 performing systems all employed deep learning algorithms. Specifically, the best system by Alibaba Inc. uses a language-level bidirectional LSTM (BiLSTM), convolutional neural network (CNN) for character-level, and CRFs for dependencies[21]. UTHealth /Dalian's system, the second-best performer in the challenge, applies a combination of BiLSTM and CRF for entity recognition[21]. Finally, word embeddings (word2vec), LSTM-CRFs, and RCNN-CRFs with a convolutional layer are the main components of UTHealth's system, which is the third-best performer[21].

## III. Evaluation Datasets

The focus of the 2009 i2b2 Medication Extraction Challenge was to extract medication information that are experienced by the patient[17]. As for the 2018 n2c2 challenge, there were different subtasks that were based on a) medication information extraction or b) adverse drug event identification[21]. Note that since ACM has currently no feature to detect adverse drug events, this work only focused on medication information extraction.

The i2b2 dataset is composed of deidentified discharge summaries from Partner's Healthcare in Boston, MA; originally, 696 clinical notes were designated as training set and 251 clinical notes were separated for unbiased testing. Only the test set was manually annotated for gold standard labels by participants and organizers [17]. On the other hand, the n2c2 corpus is a compilation of 505 deidentified discharge summaries from the MIMIC (Medical Information Mart for Intensive Care) III database[22]. Specifically, 303 notes were dedicated for model development and the remaining 202 notes were annotated for gold standard labels and were used as the official test set. Note that ACM was only evaluated on the tests sets for both the i2b2 and n2c2 challenges.

To further establish generalizability of medication extraction performance of ACM, 18 internal random outpatient notes from NYU Langone Medical Center were also included in the evaluation. An internal physician manually deidentified the clinical notes for any PHI and annotated for medication information using doccano, an open source text annotation tool[23].

The annotated medication information—which will be called "fields" from here on—for each of the test sets are described in Table 1. Examples of text notes and the corresponding ground truth labels are provided in Figure 2.

**Table 1.** Description and number of entities per field in the three evaluation datasets

| Field | Description | Number of Entities: 2009 i2b2 dataset | Number of Entities: 2018 n2c2 dataset | Number of Entities: NYU Langone Medical Center Notes |
|---|---|---|---|---|
| Name | Brand and generic names and collective names of prescription substances, over-the-counter drugs, and other biological substances. | 8495 | 26803 | 216 |
| Dosage | Amount of medication used in every single administration. | 4387 | 6900 | 134 |
| Frequency | How often each dose of medication is taken. | 3999 | 10293 | 134 |
| Mode | Method how the medication is taken into the body. | 3307 | 8987 | 138 |
| Duration | Time span for how long medication should be administered | 511 | 966 | 3 |
| Reason | Medical reason for medication intake. | 1342 | 6384 | *** |
| Strength | The amount of active ingredient in the medication per single dosage of administration. | *** | 10,922 | 99 |
| Form | Type or format the medication is presented by the manufacturer. | *** | 11,006 | 94 |

****Field was not annotated in the corresponding dataset*

---

19 …including courses of intravenous nafcillin x 4 weeks

20 and with vancomycin x 4 weeks…

m="nafcillin" 19:5 19:5||do="nm"||mo="intravenous" 19:4 19:4||f="nm"||du="x 4 weeks" 19:6 19:8||r="nm"||ln="narrative"

m="vancomycin" 20:3 20:3||do="nm"||mo="nm"||f="nm"||du="x 4 weeks" 20:4 20:6||r="nm"||ln="narrative"

A. 2009 i2b2 sample clinical text note and corresponding gold standard labels

---

…MEDICATIONS: Lipitor, Tylenol with Codeine, Dilantin, previously on Decadron…

| T1 | Drug 1094 1101 | Lipitor |
| T2 | Drug 1103 1123 | Tylenol with Codeine |
| T3 | Drug 1125 1133 | Dilantin |
| T4 | Drug 1149 1157 | Decadron |

B. 2018 n2c2 sample clinical text note and corresponding gold standard labels

---

…oxybutynin (DITROPAN) 5 mg tablet Take 5 mg by mouth 3 (three) times daily…

m= "oxybutynin (DITROPAN)" 1527 1546
do= "5 mg" 1566 1568
f = "3 (three) times daily" 1580 1597
mo= "by mouth" 1571 1577
str= "5 mg" 1547 1549
fo= "tablet" 1550 1555

C. NYU Langone Medical Center sample clinical text note and corresponding gold standard labels

**Figure 2:** Examples of clinical text notes and its corresponding gold standard labels from A) i2b2 dataset, B) n2c2 dataset, and C) NYU Langone Medical Center. There are differences in the formatting of the gold standard labels, but all follow a general format of: [field/ground truth/start offset and end offset]. In the case of A) and C), m=medication, do=dosage, f=frequency, mo=mode, du=duration, str=strength, r=reason, and fo=form. When the value of a field is "nm" in A), this implies that an entity related to that field was not mentioned in the analyzed text. No "nm" values were included in B) and C). The start offset and end offset indicate

where the entity is located in the corpus. Note that the gold standard labels for A) use the line number and the position of the entity within that line for the start offset and end offset, while gold standard labels for B) and C) utilize character offset.

**IV. Methods**

Text Preprocessing

The average length of the i2b2 corpus is 6643 characters, while the n2c2 notes are twice as long with an average of 12,550 characters. There is a 20,000 UTF-8-character limit per processed document in ACM's NERe API. This limitation posed a problem particularly for the n2c2 corpus, as there were 17 clinical text notes that exceeded 20,000 characters. To circumnavigate this caveat, these documents were subjected to text segmentation by dividing the text into smaller "blocks." This is achieved by first performing tokenization and splitting the tokenized word list into its midpoint, thus creating 2 independent text blocks. For segments that still exceeded the prescribed character length, additional midpoint splitting was employed. Ultimately, text segments that corresponded to a single document were processed in ACM's API, separately, one at a time.

Text Processing

ACM is available in multiple modes of access—web console and programmatic access via Python or Java. Experiments in this work were all performed using Python via the boto3 library to access the NERe API. Each document is processed one at a time, synchronously, and it takes approximately 11 seconds to produce structured outputs for a single document with 12,000 characters.

ACM's output includes the following: a) clinical concept (e.g. anatomy, PHI, or medication, etc.), b) type (the field where the entity belongs), c) text (actual extracted entity), d) confidence score (quantitative measure to determine how sure the system is with its prediction/extraction), e) offset (location of the extracted entity in the document), f) traits (information understood by the system based on context), and g) attributes (related information to the extracted entity)[9]. For this work, information from all other clinical concepts except medications were filtered out. The medication category provides specific information regarding rate, negation of medication administration, and all the other fields described in Table 1. Note that outputs relating to rate and negation of medication administration were also filtered out because the evaluation datasets excluded both fields in the gold standard annotations.

Text Post-Processing

For the 17 documents that were subjected to text segmentation, offset adjustments in ACM's output were necessary as the text fragments from a single document were processed separately, rather than as a single unit. For each of the text blocks, adjustments were performed by first determining the character start offset of the text block's first token in the original unsegmented narrative. This value was then added to the start and end offsets of all system-returned entities belonging to that same block.

As a simple example, consider this short note: "patient took advil for pain." There are total of 27 characters, and if this were to be segmented into text blocks with equal to or less than 14 characters in length, 2 blocks are produced: a) ["patient took"] and b) ["advil for pain]. In a), the first token's start offset ("patient") is 0 in both the text block and in the original note. Thus, when this value is added to the entity offsets in a), there are no changes in any of the numbers. This is not the case for b). Notice that "advil" is the first token, and accordingly, is designated with the start offset of 0. In the original note, however, "advil" begins as the 13$^{th}$ character. To address this discrepancy, 13 should be added to the start and end offsets of "advil" and to all the other entity offsets within b).

Medication Extraction Evaluation

Performance was measured quantitatively using a) precision (ratio of the correctly identified entities by the system and the total number of tokens extracted by the system), b) recall (ratio of the correctly identified entities by the system and the total number of gold standard labels), and c) F-score (harmonic mean of the precision and recall, where β=1). The following are their formulae:

a) Precision = True Positives / (True Positives + False Positives)
b) Recall = True Positives / (True Positives + False Negatives)
c) F-Score = ($1^2$ + 1) (Precision)(Recall) / ($1^2$ * Precision) + Recall

These metrics were employed for two types of matching: exact (also called as strict for n2c2) and inexact (also called as lenient for n2c2). Exact matches are realized at the phrase level, and this entails that the field, extracted entity, and the beginning and end offset predicted by the system must match exactly the corresponding gold standard label. Note that in the i2b2 evaluation dataset, entries that are "nm" are excluded in the exact matching calculations. Inexact matching, on the other hand, is realized at the token level (i.e. words or characters delimited by spaces and punctuations). The i2b2 challenge particularly defined that an inexact match as a partial overlap between the system-returned token and the ground truth token, while the n2c2 challenge recognized a match if the field tag is correct and the offset span partially matches with the ground truth label.

Regardless of the type of matching, the precision, recall, and F-scores were considered at two levels of granularity: document level (macro average for n2c2) and system level (micro average for n2c2). ACM's performance on the NYU clinical notes was based on system level exact matching. In the 2009 i2b2 challenge, the primary evaluation metric was also system level exact matching, while the 2018 n2c2 challenge the primary evaluation metric was micro lenient matching. F-score was the single composite evaluation metric used for system comparisons in both challenges.

Performance Comparison
ACM's performance on the i2b2 evaluation dataset and n2c2 evaluation dataset were compared against the three best performing systems in the each of the challenges. These include systems by Tao et. al, Patrick et. al., and Doan et. al and Alibaba Inc., UTHealth/Dalian, and University of Florida, respectively. The components of these systems were described in section II of this paper.

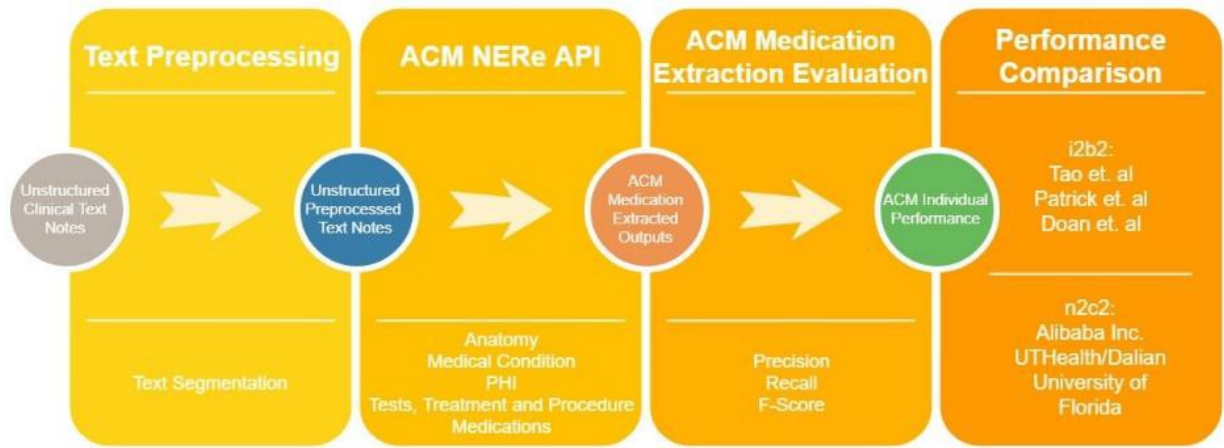

**Figure 3:** ACM evaluation methods and workflow

## V. Results
ACM Performance on the 2009 i2b2 Evaluation Dataset

**Table 2.** System level exact matching results of ACM in the 2009 i2b2 evaluation dataset

| Field | Precision | Recall | F-Score |
|---|---|---|---|
| Name | 0.794 | 0.776 | 0.785 |
| Dosage | **0.840** | **0.785** | **0.812** |
| Frequency | 0.784 | 0.673 | 0.724 |
| Mode | 0.823 | 0.744 | 0.782 |
| Duration | 0.338 | 0.135 | 0.193 |
| Reason | *** | *** | *** |
| Overall w/o Reason | 0.801 | 0.737 | 0.768 |

*** *ACM has no detect reason for medication feature*

The primary evaluation metric of the 2009 i2b2 challenge computed the system level exact matching. ACM performed noticeably the best in detecting entities in the dosage field, garnering a recall of 0.840, precision of 0.785, and F-score of 0.812. Results were also satisfactory for name, frequency, and duration fields, with recalls ranging from 0.78 to 0.82, precisions ranging from 0.67 to 0.77, and F-scores ranging from 0.72 to 0.78. However, extraction of duration entities was challenging for ACM, yielding a recall=0.338, precision=0.135, and F-score=0.193. Interestingly, systems that participated in the challenge also had low performances with the same field, with F-scores lower than 0.530[17].

Altogether, ACM achieved a system exact recall of 0.801, precision of 0.737, and F-score of 0.768 (Overall w/o Reason in Table 2). It is important to note that these values did not account for the reason entities—which was 6% of the total gold standard labels—because ACM has no detect reason feature.

**Table 3.** ACM vs. the three best systems in the 2009 i2b2 challenge

| System | F-Score |
|---|---|
| Tao et. al | **0.864** |
| Patrick et.al | 0.857 |
| Doan et. al | 0.821 |
| ACM w/o Reason Field | 0.768 |
| ACM w/ Perfect Reason Field | 0.782 |
| ACM w/ Patrick et. al's Reason Field Performance | <u>0.752</u> |

Basing from its baseline F-score without the reason field, ACM ranked the lowest as shown in Table 3. However, omission of the reason field in the evaluation of ACM presents imprecise comparisons with other systems, which accounted for this field in their final F-scores. To address this pitfall, ACM's performance was initially re-computed for a best-case scenario, assuming that the system correctly identifies the total 1342 reason entities. This assumption readjusted the overall precision to 0.810, recall to 0.746, and F-score to 0.777 (ACM w/ Perfect Reason Field in Table 3).

However, in the actual challenge, extraction of reason entities was the toughest task for all the systems in the challenge[17]. As matter of fact, the best performer for the reason field in Table 3 is Patrick et. al's system, which achieved a precision of 0.668, recall of 0.331, and F-score of 0.443. These values were utilized to derive the exact true positives, false positives, and false negatives and to be added to ACM's baseline scores for a more realistic computation and comparison. Ultimately, after the adjustment, ACM remained the lowest performer overall, garnering an F-score of 0.752 (ACM w/ Patrick et. al's Reason Field Performance in Table 3), while Tao et. al's machine learning-based system achieved the best performance with an F-score of 0.864.

ACM Performance on the 2018 n2c2 evaluation dataset

**Table 4.** Micro lenient matching results of ACM in the 2018 n2c2 evaluation dataset

| Field | Precision | Recall | F-Score |
|---|---|---|---|
| Name | 0.920 | 0.863 | 0.891 |
| Dosage | 0.481 | 0.838 | 0.612 |
| Frequency | 0.908 | **0.939** | **0.923** |
| Mode | 0.937 | 0.866 | 0.900 |
| Duration | 0.694 | 0.677 | 0.685 |
| Reason | *** | *** | *** |
| Strength | **0.983** | 0.467 | 0.633 |
| Form | 0.938 | 0.824 | 0.877 |
| Overall w/o Reason | 0.852 | 0.806 | 0.828 |

*** *ACM has no detect reason for medication feature*

Unlike the i2b2 competition, the primary evaluation metric for the n2c2 challenge was the micro lenient matching. Moreover, the n2c2 evaluation dataset incorporated the strength and form fields as part of the entity extraction task. ACM's best suit was detecting the frequency field, with a recall of 0.939 and F-score of 0.932. The system also performed well in the name, form, and mode fields, with precisions ranging from 0.908 to 0.937, recalls ranging from 0.863 to 0.939, and F-scores ranging 0.891 to 0.923. Results for dosage, duration, and strength fields were relatively lower with F-scores less than 0.70.

**Table 5**. ACM vs. the three best systems in the 2018 n2c2 challenge

| System | F-Score |
|---|---|
| Alibaba Inc. | **0.941** |
| UTHealth/Dalian | 0.934 |
| University of Florida | 0.928 |
| ACM w/o Reason | 0.828 |
| ACM w/ Perfect Reason | 0.873 |
| ACM w/ Alibaba Inc.'s Reason Performance (Recall = 0.60) | 0.827 |
| ACM w/ Alibaba Inc.'s Reason Performance (Recall = 0.70) | 0.826 |
| ACM w/ Alibaba Inc.'s Reason Performance (Recall = 0.80) | 0.825 |

The system's overall F-score without the reason entities was 0.828 (ACM w/o Reason in Table 5). As in the i2b2 evaluation, adjustments relating to the reason field were applied to the baseline F-score. First, the perfect-case scenario (ACM correctly identifies all 6384 reason entities) was assumed, which bumped ACM's F-score from 0.828 to 0.873 (ACM w/ Perfect Reason in Table 5).

For a more realistic adjustment, ACM's baseline F-score was recalculated with Alibaba Inc's reason field performance (F-score=0.728), which was the best compared to the other systems included in Table 5. Since its corresponding precision and recall for the reason field have not yet been disclosed in literature, three different scenarios were assumed: ACM correctly identifies a) 60% (Recall=0.60) of the gold standard labels, b) 70% (Recall=0.70) of the gold standard labels, and c) 80% (Recall=0.80) of the gold standard labels. With the F-score remaining as a constant in all the scenarios, the corresponding precisions were calculated. Ultimately, these equated to F-scores of 0.827 for a) 0.826 for b), and 0.825 for c). Despite all the adjustments, ACM still ranked the lowest compared to the other systems in Table 5.

## ACM Performance on the Random NYU Langone Medical Center Clinical Notes

Table 6: System level exact matching performance of ACM on random NYU Langone Medical Center clinical notes

| Field | Precision | Recall | F-Score |
|---|---|---|---|
| Name | 0.687 | 0.766 | 0.725 |
| Dosage | 0.768 | **0.906** | 0.831 |
| Frequency | **0.885** | 0.834 | **0.859** |
| Mode | 0.659 | 0.805 | 0.725 |
| Duration | 0.330 | 0.330 | 0.330 |
| Reason | *** | *** | *** |
| Strength | 0.744 | 0.646 | 0.691 |
| Form | 0.735 | 0.619 | 0.672 |
| Overall w/o Reason | 0.735 | 0.773 | 0.753 |

***ACM has no detect reason for medication feature and in this corpus, reason field was not annotated.

System level exact matching was employed to examine the quality of ACM's output in the NYU evaluation corpus. As was the case in the n2c2 evaluation dataset, ACM's strength (F-score=0.859) was extracting entities pertaining to frequency. On the other end of the spectrum, extracting duration entities was again the most challenging task for ACM (F-score=0.33), which was the case in the i2b2 evaluation. Altogether, ACM's overall precision, recall, and F-score was 0.735, 0.773, and 0.753, respectively.

## VI. Discussion

### List Entities vs Narrative Entities

Clinical text notes present medication entities via list (i.e. enumeration of medications) and/or via narratives (i.e. medications are embedded in sentences, which also often contain non-medication related information). Every single gold standard label in the i2b2 dataset included annotations indicating which text structure the entity appeared. This allowed for further analysis if ACM performs better when entities are in a list or when entities are in narratives.

Entities in list form and entities in narrative form were segregated and performance evaluation was conducted independently for each group. In total, there were 23,557 list entities and 18,553 narrative entities. Analysis showed that ACM achieved a precision=0.913, recall=0.488, and F-score=0.636 for its performance on the former and a precision=0.939, recall=0.208, and F-score=0.341 for the latter. Similarly, other systems that participated in the 2009 i2b2 challenge also scored better when extracting entities from lists than from narratives[17].

### Deficiencies and Benefits of ACM

Based on all the evaluation datasets, one of the most noticeable deficiencies of ACM pertains to misclassification of certain medications as treatment names. Some major examples that surfaced include oxygen ($O_2$), packed red blood cells (PRBCs), chemotherapy, ace inhibitor, diuretics, beta-blockers (BB), and angiotensin II receptor blockers (ARBs). Additionally, ACM had difficulties distinguishing medication names that relate to groups of medications such as home medications, cardiac medications, pain medications, narcotics, and antibiotics. While one could argue that this is a matter of interpretation, these nuances could have implications for specific use-cases, particularly when they serve as inputs to automated downstream applications.

Another curious behavior that surfaced was ACM's sensitivity to white space. This is especially applicable to medications that are composed of more than one noun. For example, if the complete medication name "polysaccharide iron complex" happens to have a new line inserted after polysaccharide, the extracted entity by ACM only included "iron complex". Given the fact that clinical notes are plagued with numerous syntactic and grammatical issues, this behavior can significantly impact performance and yield unstable results.

Currently, ACM has some technical limitations, which can be construed as deficiencies when dealing with medication information extraction. First, the system is missing the detect reason for medication feature. Information regarding medication indication are included in clinical text notes because it educates the patient, decreases inappropriate prescriptions by the healthcare provider, and overall, it improves patient safety[24]. That being said, it is crucial for an IE system such as ACM to be equipped with this feature especially when extracting medication information. Another technical limitation of ACM is the 20,000 UTF-8 document length limit. As discussed in the paper earlier, this limitation was problematic for 17 clinical text notes from the n2c2 corpus. Having information-rich and lengthy clinical notes are not uncommon across healthcare institutions in the US because aside from describing details regarding patient care, it also contains other information related to compliance and reimbursement, which ultimately cause what Kuhn et. al call as "note bloat."[25] Therefore, the current length restriction would prevent direct and immediate processing of such long notes.

Notwithstanding these shortcomings that are not unusual in a complex data product, the benefits that ACM provides are real. First and foremost, it provides a clean API that can easily scale to handle large workloads while ensuring data security and privacy. While this work focused solely on the medication extraction subtask, several other important clinical concepts can be extracted with the same API call. The underlying deep learning models and technology that power ACM have already been implemented and are likely to be updated routinely as end-user requirements evolve. Multiple modes of access—web console and programmatic access via Python or Java—cater to a wide audience with varying levels of skill and facility for technology. All this complexity is hidden from the end-users and requires only modest computing skills for effective use.

In addition, there is no upfront fee or commitment that could lock users into a singular solution, which means users only pay for what they use. This diminishes the need to invest in in-house experts for building advanced algorithms and/or managing large-scale infrastructure to support such algorithms, which is outside the mission of most hospitals and care practices, especially smaller ones.

Altogether, ACM presents a promising approach with its built-in simplicity and usability. The combination of all the above factors can empower end-users to innovate and drive usage. In terms of its performance for medication information extraction, however, better systems do exist, at least for now.

Limitations

This work has some limitations. First, ACM was only evaluated on the medication extraction task. This implies that its performance for the medication category may not necessarily reflect its performance for other clinical concepts, which include anatomy, medical condition, PHI, test name, treatment name, and procedure names. This work was also not able to analyze negation of medication entities and the field pertaining to the rate of medication because all of the three evaluation datasets were not annotated for these information. Therefore, future work should involve exploration of all these aspects. Lastly, the results presented in this work were all outputs form ACM's first release version, which was publicly available on November 27, 2018. There may have been updates and changes to its existing machinery from the time leading to the completion of this work on June 2019.